\documentclass[runningheads]{llncs}

\usepackage{eccv}

\usepackage{eccvabbrv}

\usepackage{graphicx}
\usepackage{booktabs}
\usepackage{amsmath}
\usepackage{amssymb}

\usepackage[accsupp]{axessibility}  %

\usepackage[hidelinks]{hyperref}

\usepackage{orcidlink}
\usepackage{needspace}

\begin{document}

\title{Native Association: Confidence-Aware Human Perception
in the Wild with a Foundation VLM}

\titlerunning{Native Association: Human Perception in the Wild}

\author{Igal Dmitriev \and Ofir Liba}
\authorrunning{I.~Dmitriev and O.~Liba}
\institute{WSC Sports, Tel Aviv, Israel\\
\email{\{igal.dmitriev,ofir.liba\}@wsc-sports.com}}

\maketitle

\begin{abstract}
Extracting who is where, on which team, wearing which number from a broadcast
frame is typically done by stitching a detector, an OCR engine, and classifiers
together---and the stitching step swaps identities under occlusion. We make
association \emph{native} instead: a 0.77B vision--language model (Florence-2) is
fine-tuned to emit all per-person attributes as one grammar-constrained sequence,
with each attribute generated inside its owner's block. Output is therefore
schema-valid on every frame by construction, and no post-hoc binding step exists
to attach a correctly read number to the wrong player: residual misassociation is
pure perception error, $\approx4\times$ rarer than zero-shot-prompted frontier APIs' ($0.057$ vs.\
$0.21$--$0.24$). On a frozen multi-sport test set, this single pass reaches 0.95
detection F1 (APIs: 0.65--0.75). A single extra forward pass yields a per-field
confidence that supports a reject option (jersey precision $0.71\rightarrow0.96$
at half coverage) and routes a training-free zoom-and-re-read for small players.
Surprisingly, once the grammar is learned, further parameter-efficient tuning
yields no measurable gain under the adaptation configurations we test; the
identical recipe on WIDER-Attribute reaches 93.1 mAP given-box, yields the
first detection-coupled end-to-end results under its standard test protocol
(84.5 mAP), and reproduces the same tuning result.
In this regime, the gains live in the structure, not in added weights.
\end{abstract}

\begin{figure}[!t]
  \centering
  \includegraphics[width=0.86\textwidth]{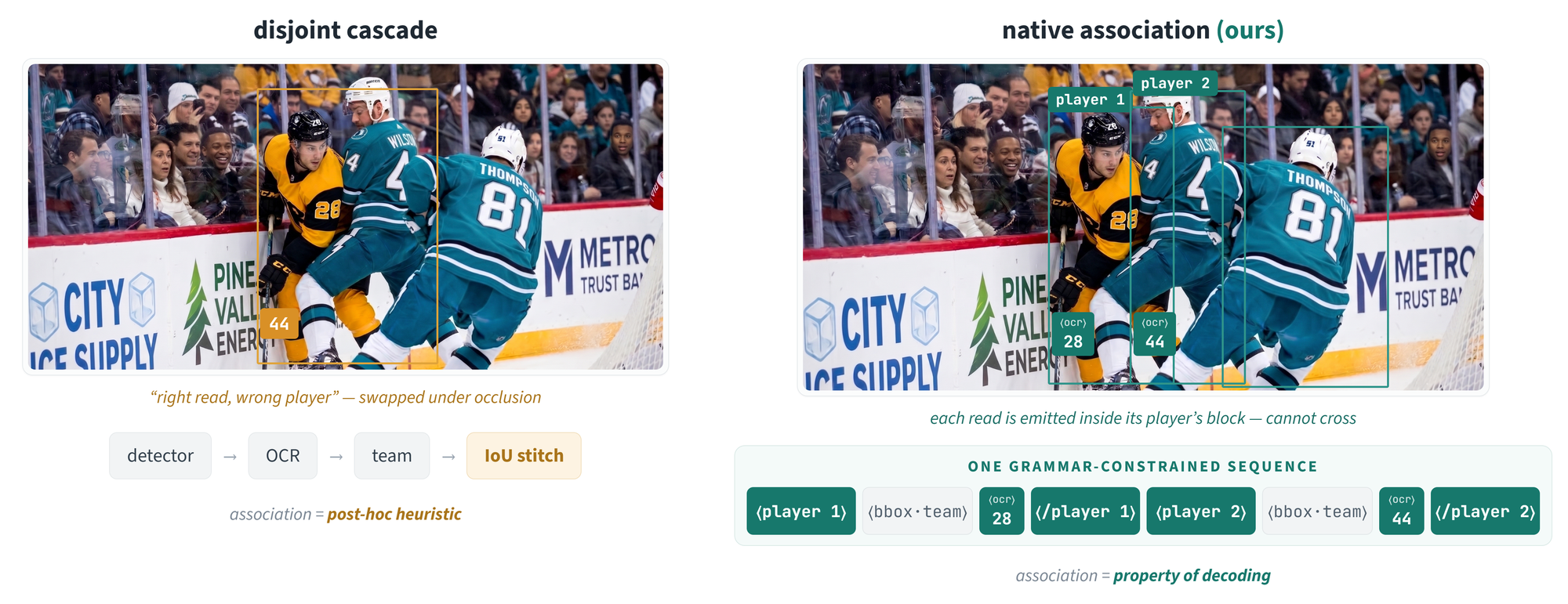}
  \caption{Native association vs.\ post-hoc stitching. A disjoint cascade reads
    the number correctly but binds it to the wrong athlete via IoU stitching under
    occlusion (left); our model emits each read inside its owner's
    \texttt{<player>} block of one grammar-constrained sequence, so the binding
    step does not exist (right). Quantified
    in \cref{tab:main} (jersey-digit AER) and \cref{sec:exp-robust}.}
  \label{fig:teaser}
\end{figure}

\section{Introduction}
\label{sec:intro}

Dense, grounded, per-person metadata---every athlete's box, team, jersey number,
on-body text, and scene context---is a deployment problem before it is a modeling
problem: it underpins sports analytics and archive-scale retrieval. Sports
broadcast frames are a concrete instance of human-centered perception in the wild:
motion blur, heavy inter-player occlusion, extreme scale variation (one athlete
may fill the frame, a teammate ${<}128$ pixels), and small, rotated, warped
typography.

Two families dominate deployment, and both fail at the same joint: association.
\emph{Disjoint cascades} chain a detector, an OCR engine, and classifiers, then
stitch their outputs by an IoU heuristic; under occlusion the stitch attaches a
correctly read number to the wrong athlete---even granted ground-truth boxes
and one crop per athlete, cascades still misbind $15$--$18\%$ of the jersey
digits they read (supplementary).
\emph{Frontier vision--language APIs} offer strong open-world semantics but---on
this task---weak grounding, unreliable output structure, no per-field confidence,
and seconds of latency per frame (\cref{sec:exp-robust}).

These failures live on three distinct axes, kept separate throughout:
\emph{syntactic validity} (does the output parse into the schema on every
frame?), \emph{visual grounding} (is every attribute anchored to a real,
non-degenerate region?), and \emph{semantic association} (is every attribute
bound to the correct person?). Cascades fail chiefly the third axis (the
stitch); the APIs degrade on all three.

We make association native instead, mapping one mechanism to each axis. A single
0.77B foundation VLM is fine-tuned to emit all per-person attributes as one
grammar-constrained sequence, each athlete's number and text generated
\emph{inside} that athlete's block (\cref{fig:teaser}). The finite-state grammar
makes every frame schema-valid by construction (validity); in-block emission
removes the post-hoc binding step, cutting the jersey-digit association error
rate (AER: the fraction of correctly read numbers not bound to their true owner;
\cref{sec:data}) to about a quarter of the APIs' (association); and, because
grounding quality is pixel-bound for small players, a confidence-routed,
training-free crop re-query re-reads exactly them (grounding), with one extra
teacher-forced pass exposing the per-field confidence that routes it and supports
a reject option (\cref{sec:exp-conf}).

To test that this is a recipe rather than a lucky checkpoint, we transplant it
unchanged---same schema, grammar, and frozen towers---to the public
WIDER-Attribute benchmark~\cite{widerattribute} (\cref{sec:wider}). On both
domains, further parameter-efficient tuning beyond the fine-tuned decoder moves
the deployed metric by less than noise under the configurations we test
(\cref{sec:exp-abl}); in this regime the gains live in the structure, not in
added weights. All
systems share one frozen, fingerprinted test set and size-stratified analysis
(\cref{sec:exp-size}).

\Needspace*{6\baselineskip}
\subsubsection{Contributions.}
\begin{itemize}
  \item \textbf{A native-association single-pass architecture.} Grammar-constrained
  decoding emits every attribute \emph{inside} its owner's block, eliminating the
  post-hoc binding step---${\approx}4\times$ less misassociation than frontier
  APIs---and guaranteeing schema-valid output on every frame (\cref{sec:exp-robust}).
  \item \textbf{Deployable per-field confidence.} One teacher-forced replay yields
  per-field confidence---a signal no baseline exposes---supporting a reject option
  (jersey precision $0.71\rightarrow0.96$ at half coverage) and routing a crop
  re-query.
  \item \textbf{No measurable gain from tuning beyond the grammar.} Once the
  grammar is learned, further parameter-efficient tuning beyond the fine-tuned
  decoder shows no measurable gain under the tested adaptation configurations,
  on both domains
  (\cref{sec:exp-abl}); the same recipe reaches 93.1 mAP given-box and 84.5 mAP
  end-to-end on WIDER-Attribute (to our knowledge the first detection-coupled
  result), with checkpoint and recipe released (\cref{sec:wider}).
\end{itemize}

\section{Related Work}
\label{sec:related}

\subsubsection{Autoregressive spatial perception.} Recasting localization as
sequence generation is established---Pix2Seq emits boxes as discrete
tokens~\cite{pix2seq}, Kosmos-2 grounds phrases to regions~\cite{kosmos2}, and
Florence-2 unifies detection, captioning, and OCR in one prompt-conditioned
decoder over a DaViT backbone~\cite{davit,florence2}---as is constrained
decoding for entity linking~\cite{genre}, text-to-SQL~\cite{picard}, and
general grammar-constrained LLMs~\cite{outlines}. Zero-shot, these AR models
cannot serve as comparison rows: neither Pix2Seq nor Kosmos-2 emits a
schema-typed, multi-field per-person record (team, jersey, on-body text), so no
digit-association protocol exists to score them under; fine-tuning them on our
corpus would change the backbone, not the mechanism under test. DETR-style
object queries~\cite{detr} bind attributes to instances natively but require
trained task heads per attribute and offer no schema- or text-validity
guarantee;
generative grounded-NER frameworks~\cite{gmner} emit entity--region
associations within one sequence, but over image--text pairs with
detector-proposed regions and no validity guarantee, and multimodal constrained
decoding has targeted single-entity outputs~\cite{autover}. Our contribution is
thus not any single component---structured serialization, constrained decoding,
token-likelihood confidence, and crop re-querying are each established---but
the composition in which the grammar \emph{carries association}, not merely
well-formedness: to our knowledge, we are the first to use a finite-state
grammar to enforce association \emph{and} schema validity by construction in
multi-entity extraction from images alone.

\subsubsection{Frontier VLMs and task-specific recognition.} Production
vision--language APIs (\eg, Gemini~\cite{gemini}) can be prompted to return
per-person JSON and, in schema-constrained modes, guarantee JSON
well-formedness; no mode, however, guarantees grounded, non-degenerate geometry
or exposes per-field confidence, at seconds per frame. We evaluate the APIs
free-form---their accuracy-best configuration~\cite{letmespeak}---and report
first-attempt validity in \cref{sec:exp-robust}; our decoder guarantees
schema-valid, non-degenerate output with a usable confidence at
$6.9$--$21\times$ lower latency. Person/pedestrian attribute recognition (PAR)
targets the same per-person labeling but on \emph{given} crops, from
attribute--identity CNNs~\cite{apr} to LLM-based recognizers~\cite{llmpar}
(no comparable end-to-end WIDER number); none predict attributes jointly with
their grounding, as our single pass does.

\subsubsection{Disjoint pipelines and confidence.} The dominant deployed
approach stitches specialists---a detector~\cite{groundingdino,sam}, an OCR
engine~\cite{ppocr,trocr,donut}, and a team/attribute classifier---by post-hoc
IoU~\cite{soccernetgsr}; each component can be strong in isolation, but the
stitch is fragile under occlusion. SoccerNet jersey-number
systems~\cite{koshkina2024jersey,grad2025jersey} read numbers from tracklet
crops with detection assumed---not directly comparable to our per-frame
detect-and-bind protocol. Calibration~\cite{calibration} and reject-option
classifiers~\cite{selective} trade coverage for reliability but are rarely
available from generative VLMs; we obtain per-field confidence from one
teacher-forced replay, show it yields deployable reject options on both sports
and WIDER-Attribute~\cite{widerattribute}, and report which internal signals
are uninformative.

\section{Method}
\label{sec:method}

Our engine is a single foundation vision--language model (Florence-2-large,
0.77B parameters~\cite{florence2}) fine-tuned to map one image to one
structured sequence that names every athlete and all of their attributes.
Florence-2 couples a DaViT vision encoder to a BART-style encoder--decoder:
visual embeddings and the embedded task prompt are fused by the transformer
encoder, and our serialization is the decoder target. We freeze both encoders
and fine-tune the decoder and shared token embeddings (details in the
supplementary). The fine-tuning aligns the vocabulary; the guarantees come from
the structure: a fixed-slot schema and finite-state grammar make association
structural and schema validity guaranteed (\cref{sec:method-grammar}), one replay pass exposes per-field confidence
(\cref{sec:method-conf}), and an optional confidence-routed crop re-query
recovers small players without training (\cref{sec:method-requery}). The
training recipe is deliberately minimal (\cref{sec:method-train}).

\subsection{Unified Schema and Grounded Single-Pass Decoding}
\label{sec:method-grammar}

We expand the tokenizer with atomic structural tokens
(\texttt{<sport>}, \texttt{<stype>}, \texttt{<player>}, \texttt{<bbox>},
\texttt{<team>} and its four flags, \texttt{<ocr>}, \texttt{<gdesc>}) and
serialize the full per-frame annotation into one target sequence whose emission
order is enforced by a prefix-constrained finite-state grammar---the
serialization below is the only shape a decoded sequence can take (visualized in
the supplementary):
\begin{center}
\begin{minipage}{\linewidth}\centering\footnotesize\ttfamily
<sport>\{s\} <stype>\{t\}\\
\ \ ( <player> <bbox> <loc>$\times$4 </bbox> <team>\{flag\}\\
\ \ \ \ \ \ ( <ocr>\{text\} [<loc>$\times$8] )*\\
\ \ \ \ </player> )*\\
<gdesc>\{desc\} </s>
\end{minipage}
\end{center}
Continuous coordinates are quantized to 1000 location bins and emitted as
\texttt{<loc\_*>} tokens; the implemented grammar bounds the starred repetitions
(12 players, 17 OCR items per player). Player blocks are emitted largest-first
(bbox area, descending) and OCR items largest-quad-first---a deterministic
canonical order; under the grammar we observe none of the repetition or
early-termination failures that motivated Pix2Seq's randomized ordering and
sequence augmentation~\cite{pix2seq} (removing the grammar, by contrast,
collapses decoding entirely; \cref{sec:exp-abl}). The eight per-OCR quad tokens
(bracketed above) are optional: our headline configuration omits them, since
quad supervision is net-negative for reading ($-3.8$ jersey-F1 for $+0.4$
detection-F1; ablation in \cref{sec:exp-abl}); the quad variant is retained for
downstream tasks needing per-word geometry.

Two properties follow \emph{by construction}, not by training. First,
\textbf{native association}: every OCR string (and, when emitted, its
coordinates) is generated inside exactly one
\texttt{<player>}\dots\texttt{</player>} block, so text--player binding is read
off the sequence rather than reconstructed by a post-hoc IoU stitch. This does
not make the binding infallible: a residual \emph{perception} error remains, in
which the model reads a neighbour's number into the wrong block; we measure it
as the association error rate (AER; defined in \cref{sec:data})---$0.057$ single-pass,
$\approx4\times$ below the frontier APIs' $0.21$--$0.24$, whose grounding is
itself post-hoc. Second, \textbf{guaranteed validity}: a prefix-allowed-tokens
function admits only grammar-legal continuations at each step, so every decoded
sequence parses into the schema---validity $1.0$ on every frame, with no
post-hoc repair and no reliance on the model having ``learned'' to be
well-formed. The result is a \emph{typed serialization} whose single-token keys
avoid literal JSON's 3--5 BPE tokens per key and its malformability.

\subsection{Per-Field Confidence from a Single Teacher-Forced Replay}
\label{sec:method-conf}

After constrained generation, one teacher-forced forward pass (fp32) over the
generated tokens recovers the per-token log-probabilities---no second
generation, no auxiliary head ($\approx150$\,ms, $\approx10\%$ of generation;
supplementary). Generation-time scores are grammar-masked, so we replay
\emph{unmasked}: confidences are full-vocabulary probabilities---the model's raw
beliefs, not values renormalized over the few grammar-legal tokens. For any
contiguous field $F=(y_{t_1},\dots,y_{t_k})$ we define its confidence as the
geometric mean of its per-token probabilities,
\begin{equation}
  c(F) \;=\; \exp\!\left(\frac{1}{k}\sum_{i=1}^{k}
  \log P\!\left(y_{t_i}\mid y_{<t_i}, X\right)\right),
  \label{eq:conf}
\end{equation}
the length-normalized likelihood of the field given the image $X$ (for
one--two-token jersey numbers, close to a single-token probability; an ablation
of the aggregator---geometric vs.\ arithmetic mean vs.\ min-probability over the
same replay logits---is in the supplementary, and finds the choice immaterial
for these short fields). Applied to
the text tokens of an OCR item, \cref{eq:conf} gives its $\mathit{text\_conf}$;
a player's $\mathit{jersey\_conf}$ is the $\mathit{text\_conf}$ of its first
digit-only OCR item (an emission-order tie-break; largest-quad when quads are
emitted). From the same logits we also read a structural $\mathit{loc\_conf}$
(geometric mean over the four box-location tokens) and an \emph{objectness}
margin ($P(\texttt{<player>})$ vs.\ $P(\texttt{<gdesc>})$, renormalized over
that two-way choice at the block-opening step). \cref{sec:exp} shows
$\mathit{jersey\_conf}$ separates correct from incorrect reads, supporting a
reject option, while the structural signals saturate and serve only to route
the re-query---a negative result we report explicitly. Cascade components
expose per-component scores (detector objectness, OCR recognition confidence),
but no unified confidence over the associated player--number pair; closed VLM
APIs expose none by default.

\subsection{Training-Free Crop Re-Query}
\label{sec:method-requery}

Small and heavily occluded athletes are the dominant error mode
(\cref{sec:exp}); we address them at inference time only, with a
zoom-and-re-read of the selected players. From the first-pass prediction we
select players that are small (bottom quartile by predicted box area) or
uncertain (bottom quartile by $\mathit{jersey\_conf}$), crop a padded region
around each, and re-run the \emph{same} model on the crop under the same
grammar capped at one player. The re-read attributes replace the first-pass
ones in place---boxes are never re-localized, so no duplication can arise---and
each emitted player is tagged with its source (frame or crop). The trade is
explicit: re-reading recovers small players ($+0.14$ tiny-bucket jersey F1) at
a small association cost ($0.057\to0.068$ AER; \cref{sec:exp-robust}), since a
crop that straddles a pile-up can bind a neighbour's number. No test label
informs the selection; the padding fraction is chosen on validation by
jersey-number F1, and crops decode as short single-player sequences, so the
average cost stays modest (1.93 passes/image; \cref{sec:exp}). The
confidence-gated flow is diagrammed in the supplementary.

\subsection{Training Recipe}
\label{sec:method-train}

Training is intentionally plain. We warm-start the new token rows with the mean
embedding of descriptive sub-words (\eg,
\texttt{<player>}~$\leftarrow$~``\,player''), freeze both encoders, and
fine-tune the decoder plus the shared token embeddings with a uniform
cross-entropy loss; the grammar plays no role in the teacher-forced
loss---validation decoding for checkpoint selection is grammar-constrained from
the first epoch. Checkpoints are selected on a held-out split by an aggregate
$\mathit{val\_score}$ (unweighted mean of five task metrics; components in the
supplementary); differences below $+0.005$ on this 0--1 composite are treated
as noise (single-seed runs; see limitations), and the frozen, fingerprinted
test set is touched once. We use no multi-stage curriculum, loss weighting, or
adapters in the main model; the adapter arm is quantified as a null in
\cref{sec:exp-abl} (full recipe and hyperparameters in the supplementary).

\section{Dataset and Evaluation Protocol}
\label{sec:data}

\subsubsection{Data and splits.}
Our sports corpus is broadcast frames from multiple sports, deduplicated
(perceptual hash) and quality-filtered. Each frame carries per-person
annotations---axis-aligned box, team $\in$ \{A, B, other, unknown\}, jersey
number (digit string, possibly empty), on-body OCR items (text, optional
quadrilateral)---plus sport type, scene type, and a free-form description. The
corpus spans five sports---hockey, soccer, gridiron football, tennis, and
basketball, plus a small \emph{other} bucket (per-sport frame and player counts
in the supplementary)---is dominated by in-game footage, and was labeled by an
LLM-assisted pre-annotation pass corrected by human annotators in CVAT over
iterative review cycles.
One frame-level partition gives train/val/test = 4{,}584/577/567 frames
(10{,}408/1{,}325/1{,}266 players; 33{,}633/4{,}185/4{,}046 OCR items).
The split is keyed on each frame's perceptual hash (stratified by sport and
athlete count), so identical or near-duplicate frames cannot straddle splits.
Provenance metadata shows 5{,}623 of the 5{,}728 frames are unique-URL still
images; 105 come from 22 source videos, of which three span splits: exactly
4/567 test and 6/577 validation frames share a video with training. Match
identity across \emph{distinct} still assets is untracked, so photographs of
one match may fall in different splits; we read our numbers as within-domain
and leave a match-disjoint split to future work.
Validation drives all model and hyper-parameter selection; the test split is
frozen and decoded once per model variant---every subsequent analysis is an
offline re-score of those cached predictions---under a fingerprint guard: each
evaluation recomputes the frame count, frame-set hash, and annotation-content
hash and aborts on mismatch (provenance in the
supplementary).

\subsubsection{Metrics.}
One evaluation suite scores every system---ours, APIs, cascades---with identical
matching. \emph{Detection}: Hungarian assignment at IoU~$\ge0.5$ (P/R/F1).
\emph{Team}: exact accuracy (which scores the annotator's A/B convention) and a
permutation-invariant purity---per-frame Hungarian assignment between predicted
and ground-truth team labels, micro-averaged over matched players.
\emph{Jersey} (digits only): the any-OCR rule---a ground-truth number is correct
if it appears as any emitted digit token for the matched player (strict
exact-string variant in \cref{sec:exp-abl}). \emph{OCR text}: fuzzy Hungarian at
normalized Levenshtein $\tau=0.7$, pooled per frame and deliberately
association-blind---binding errors are charged to AER, not text F1.
\emph{Output validity}: parse-success, empty-output, schema-valid rates.
\emph{Confidence} (\cref{sec:exp-conf}): AUROC, ECE, precision at fixed coverage.
Scene metrics and full definitions are in the supplementary. \emph{Association}:
AER is the fraction of correctly-read jersey digits not bound to their true
owner---covering both binding to a wrong matched player and grounding to no valid
box (unmatched or degenerate)---with the same matcher for every system. We report
the digit-only rate because team text is identical across teammates, so its owner
is ill-defined; we computed the all-text rate first, and its full decomposition
(all systems, all error classes) is in the supplementary. The choice matters:
under the all-text rate the oracle-boxed cascades attain \emph{lower} AER than
our single pass ($0.057$--$0.070$ vs.\ $0.085$), driven by exactly those
owner-ambiguous team-text items; the digit-only ranking is specific to it.
For oracle-boxed
cascades the denominator spans only the numbers they read
(\cref{tab:main}$^\dagger$).

\subsubsection{Size stratification.}
To expose the small-player regime, IoU-matched players are bucketed into
quartiles by ground-truth box area; detection recall, jersey F1, and OCR-text
F1 are reported per bucket (\cref{sec:exp-size}).

\subsubsection{Latency and cost.}
Accuracy metrics are computed in \texttt{fp32}; self-hosted latency is an
\texttt{fp16} pass over the full test set (the grammar ablation in the
supplementary reports \texttt{fp32}); API latencies are end-to-end
wall-clock and not strictly cross-comparable (details in the supplementary).

\subsubsection{Reproducibility.}
We will release: the evaluation suite, the finite-state
grammar and inference harness, the verbatim API prompts and retry policy, the
per-system prediction caches behind every number in \cref{sec:exp}, the
WIDER-Attribute checkpoint, and a recipe that reproduces the training and full
pipeline on any similarly-annotated dataset. The sports corpus and checkpoint are
proprietary and not released: the WIDER-Attribute results are thus reproducible
end-to-end; the sports numbers are re-scorable from the released caches, whose
provenance is pinned by the frozen-test fingerprint.

\section{Experiments}
\label{sec:exp}

\subsection{Main Comparison}
\label{sec:exp-main}

We benchmark against two production vision--language APIs (Gemini-2.5/3.1-pro)
and two disjoint tiers---an out-of-the-box Florence-2 cascade and a specialist
detector--OCR--classifier pipeline---on the frozen 567-frame multi-sport test
set, all scored through the \emph{same} suite with identical matching
(Hungarian, IoU~$\ge0.5$) so the columns are directly comparable
(\cref{tab:main}).

Protocol notes for \cref{tab:main}. Jersey metrics use the any-OCR digit rule,
with the strict exact-digit variant alongside (\cref{sec:data}). Gemini-3.1-pro
is its strongest (medium-resolution) configuration; the \emph{specialist
cascade} is a PP-OCRv5 reader, and the two cascades share one team classifier.
Match coverage is the fraction of GT players IoU-matched (equals detection
recall). First-attempt parse failure counts first-try invalid JSON: other API
cells are post-retry, ours is zero by construction, and Gemini-2.5-pro logged no
retry telemetry. Latency and cost follow \cref{sec:data}: \texttt{fp16}
self-hosted (\$1.5--2/hr on-demand A100) vs.\ wall-clock list-price for the
APIs---not cross-comparable.

\begin{table}[t]
  \caption{Main comparison on the frozen 567-frame test set (1{,}266 players,
    4{,}046 OCR items); all systems scored by the identical suite (metric rules
    in \cref{sec:data}, protocol notes in \cref{sec:exp-main}).
    \emph{Ours}~(E1b)~$=$~single pass; \emph{+re-query}~(E3)~$=$~the
    training-free crop re-query (\cref{sec:method-requery}); these labels name
    the two configurations throughout. Our AER is statistically tied with the
    oracle-boxed cascades', and the single-pass jersey-F1 lead over Gemini-3.1 is
    within CI (\cref{sec:exp-robust}).
    $^\dagger$Oracle GT boxes---generous to the cascades; their AER spans only the
    $32$--$37\%$ of numbers they read (supplementary).
    $^\ddagger$Reading$+$team$+$stitch only, \texttt{fp32}, oracle boxes---no
    detector was run, so this latency (and its cost) is not end-to-end. All
    systems run one pass per image except \emph{+re-query} ($1.93$ on
    average). Bold marks
    the best directly-comparable cell per row; oracle-assisted ($\dagger$),
    non-comparable ($\ddagger$), and inapplicable (\texttt{--}) cells are
    excluded. $\downarrow$~$=$~lower is better; \texttt{--}~$=$~not applicable.}
  \label{tab:main}
  \centering
  \scriptsize
  \setlength{\tabcolsep}{3pt}
  \begin{tabular}{@{}lcccccc@{}}
    \toprule
    & Ours & Ours & Gemini & Gemini & Specialist & Florence-2 \\
    Metric & (E1b) & +re-query & 2.5-pro & 3.1-pro & cascade$^\dagger$ & cascade$^\dagger$ \\
    \midrule
    Detection F1                     & \textbf{0.950} & \textbf{0.950} & 0.652 & 0.745 & GT$^\dagger$ & GT$^\dagger$ \\
    Team acc.\ (exact)               & 0.809 & 0.809 & 0.838 & \textbf{0.880} & 0.558 & 0.558 \\
    Team purity (perm-inv)           & 0.952 & 0.952 & 0.974 & \textbf{0.994} & 0.764 & 0.764 \\
    Match coverage                   & \textbf{0.950} & \textbf{0.950} & 0.704 & 0.772 & GT$^\dagger$ & GT$^\dagger$ \\
    Jersey F1                        & 0.757 & \textbf{0.792} & 0.704 & 0.737 & 0.460 & 0.520 \\
    Jersey F1 (strict)               & 0.695 & \textbf{0.733} & 0.630 & 0.673 & 0.407 & 0.478 \\
    Jersey P                         & 0.755 & 0.786 & 0.829 & \textbf{0.860} & 0.809 & 0.852 \\
    Jersey R                         & 0.759 & \textbf{0.799} & 0.612 & 0.645 & 0.321 & 0.374 \\
    OCR-text F1                      & 0.584 & \textbf{0.602} & 0.476 & 0.502 & 0.283 & 0.348 \\
    AER (jersey digits)$\downarrow$  & \textbf{0.057} & 0.068 & 0.242 & 0.214 & 0.075$^\dagger$ & 0.060$^\dagger$ \\
    First-attempt parse fail.\ (\%)$\downarrow$ & \textbf{0.0} & \textbf{0.0} & -- & 10.1 & 0.0 & 0.0 \\
    Schema valid (\%)                & \textbf{100} & \textbf{100} & -- & -- & -- & -- \\
    Latency (ms/frame)$\downarrow$   & 1224 & 1274 & 26009 & 8426 & 89$^\ddagger$ & 708$^\ddagger$ \\
    Cost (\$/1k)$\downarrow$         & \multicolumn{2}{c}{$\sim$0.5--0.7 (A100)} & 9.11 & 7.11 & $\sim$0.05$^\ddagger$ & $\sim$0.4$^\ddagger$ \\
  \bottomrule
  \end{tabular}
\end{table}

Three patterns summarize \cref{tab:main}. \emph{Grounding is the chasm}: we detect
at F1 $0.95$ where the APIs reach $0.65$--$0.75$, and their reading deficit
disappears once detection is conditioned away
(given-detected jersey recall: APIs $0.82$--$0.83$, ours $0.78$;
supplementary)---it is a detection deficit.
\emph{Precision-by-abstention}: every baseline family reads only the easy players
(coverage $0.70$--$0.77$; oracle cascades read $32$--$37\%$ of numbers), which
inflates their precision; we read everyone and can trade coverage for precision
explicitly (\cref{sec:exp-conf}). At each API's own recall ($0.65$ / $0.61$) our
reject option reaches jersey precision $0.91$ / $0.92$ (E1b) vs.\ their
$0.86$ / $0.83$ (supplementary): we dominate the precision--recall trade and,
unlike the APIs, can set this operating point explicitly. \emph{Association}: our misbinding rate is
$\approx4\times$ lower than the APIs' (\cref{sec:exp-robust}); a qualitative
comparison on a crowded frame is in the supplementary. One scoping note: \cref{tab:main} compares the complete
fine-tuned system with zero-shot baselines---deployment-level superiority, not
a per-component decomposition; the oracle-boxed cascade rows isolate the
stitching step (\cref{sec:exp-robust}), and a domain-trained cascade, which
would separate native association from domain adaptation, is future work
(\cref{sec:conclusion}).

\subsection{Performance by Player Size}
\label{sec:exp-size}

Reading is size-bound, and detection is where the APIs lose small players (full
size table and plot in the supplementary). For every system, jersey and OCR-text
F1 rise with player size---monotonically except in two cells---so the residual
error budget lives in the small-player regime. The failure modes differ, however: our tiny-bucket
detection recall is $0.85$ vs.\ Gemini-3.1's $0.56$---an API that never
localizes a small athlete cannot read it---whereas our own tiny-bucket
shortfall is a \emph{reading} deficit on \emph{detected} players. The
training-free crop re-query (\cref{sec:method-requery}) targets exactly this
bucket, lifting tiny-bucket jersey F1 $0.61\rightarrow0.75$ ($63$ reads gained,
$26$ lost, net $+37$) while leaving larger buckets unchanged.

\subsection{Ablations}
\label{sec:exp-abl}

Our central observation is that, once the grammar is learned, accuracy is governed by
inference-time and schema \emph{structure}, not by additional parameter tuning.
Parameter-efficient adapters (LoRA and relatives~\cite{lora}) are the default
recipe for squeezing extra accuracy from a frozen backbone, so we include a
LoRA-target ladder as an ablation and find it null---a negative result that
motivates this framing rather than a new tuning method.
Three ablations on the frozen test (all at the pre-registered
\texttt{best\_model\_score} checkpoint, jersey scored by the strict
exact-digit rule for internal consistency) support this.

\subsubsection{Quad supervision (a structural choice).}
Emitting the eight per-OCR quadrilateral tokens is optional
(\cref{sec:method-grammar}). Dropping them (our headline E1b) \emph{improves}
reading and speed while costing almost nothing on detection
(full table in the supplementary): $+3.0$ jersey-F1 under the strict exact-digit rule
($+3.8$ under the any-OCR digit-token rule used in \cref{tab:main}), $+1.6$
OCR-text-F1, $+3.7$ team purity, and a $28\%$ latency reduction
($1710\!\to\!1224$ ms), against a $-0.4$ detection-F1 change. The extra geometry
supervision spends decoder capacity that is better used for reading; we keep the
quad variant only for the ablation and for downstream tasks that need per-word
boxes.

\subsubsection{The LoRA ladder shows no measurable gain.}
Starting from the trained decoder we add LoRA adapters of increasing capacity and
re-select on validation: no adapter variant (decoder~V / vision / decoder~QVO /
decoder~QKVO, up to $3.15$M trainable parameters) moves the aggregate by more than
$+0.0011$, roughly $5\times$ under the $+0.005$ floor
(\cref{sec:method-train}; table in the supplementary). These are single-seed
runs gated by a pre-registered noise floor, so the claim is scoped accordingly:
no measurable gain was observed under the tested adaptation configurations, not
that further tuning is universally ineffective. For contrast, the single
\emph{structural} decision of dropping quads (above) is worth $+0.008$
aggregate---roughly $7\times$ the best tuning delta---in this regime (frozen
towers, a 0.77B backbone, and our data scale), structural choices rather than
added capacity move the metric, and the pattern replicates on WIDER-Attribute
(analogous tuning delta $+0.02$ mAP,
\cref{sec:wider}).

\subsubsection{The grammar constraint is load-bearing.}
In both domains, decoding the identical \texttt{best\_model\_score}
checkpoint with the same greedy configuration but the finite-state constraint
disabled collapses generation into an unbounded start-token loop: the decoder
never emits a content token, hits the $1024$-token cap, and yields zero
parseable persons (detection and jersey F1 $0.000$, empty-output rate $1.000$,
$7\times$ the latency)---on WIDER-Attribute exactly as on sports (full table
and mechanism in the supplementary). The grammar's first-token mask forbids
this continuation, so the constraint both guarantees the schema \emph{and}
terminates generation (on sports, reducing wall-clock by $7{,}580$\,ms/frame,
$-85.6\%$). Read narrowly: this ablation isolates the grammar's contribution to
\emph{decoding stability} and \emph{format validity} only---no content token is
ever emitted, so it says nothing about perception accuracy and we do not
interpret it as a detection or recognition gain. Nor do we claim the loop is
irrecoverable by other decode-time heuristics; the grammar removes it while
\emph{simultaneously} guaranteeing schema validity by construction, which
ad-hoc anti-repetition fixes do not.

\subsection{Confidence and the Reject Option}
\label{sec:exp-conf}

The single teacher-forced replay (\cref{sec:method-conf}) yields a per-player
$\mathit{jersey\_conf}$ at no extra generation cost that separates correct from
incorrect reads well (AUROC \textbf{0.879} E1b, $0.831$ after crop re-query E3;
team-margin AUROC $0.84$; ROC and distributions in the supplementary). Because it
is a length-normalized likelihood in $[0,1]$, ranking players by it and abstaining
on the least-confident trades coverage for precision (risk--coverage curve and operating
points in the supplementary): E1b reads at precision $0.71$ at full coverage, $0.87$
at the top $70\%$, and $0.96$ at the top half---a knob neither the cascades nor the
closed APIs offer, with the E1b and E3 curves crossing near full coverage. Crucially
this is opt-in---our headline numbers are at full coverage, so the reject option adds
precision without quietly dropping recall to obtain the \cref{tab:main} figures---and
it relies only on the \emph{ordering} of $\mathit{jersey\_conf}$, not its calibration
(the raw likelihood is over-confident, ECE $0.12$--$0.15$ across E1b--E3, removable with one
temperature if absolute probabilities were needed).

\subsubsection{Negative findings.}
Not every signal is informative: the structural \texttt{loc\_conf} (AUROC
$0.507$ E1b, $0.573$ E3) and the objectness margin---a renormalized two-way
choice by construction, consistent with its saturation---are near-random
correctness predictors on both domains, useful only as re-query routers
(\cref{sec:method-requery}), never filters (detail in the supplementary).

\subsection{Robustness and Statistical Reliability}
\label{sec:exp-robust}

\subsubsection{Reliability, association, and significance.}
Grammar-constrained decoding parses on all $567$ frames; the APIs carry no such
guarantee (first-attempt failures in \cref{tab:main}; full retry and
empty-output accounting in the supplementary). The damaging failure for a
grounding task is \emph{invalid geometry}: Gemini-3.1-pro emits a degenerate
(zero-area) bounding box\footnote{Zero width or height after normalization
($w\le0.002$ or $h\le0.002$); such a box has $\mathrm{IoU}=0$ with any
ground-truth player and can be matched to no one.} on $20.3\%$ of its predicted
players---at least one on $22.2\%$ of frames---and on $177$ of those players
the jersey text was \emph{right} but grounded to no athlete, so it cannot be
credited: reading without grounding, exactly the failure in-block emission
removes. Consistently, $90$--$97\%$ of API misbindings are such orphaned reads,
whereas ours are contact-occlusion swaps between overlapping players
(decomposition in the supplementary); and our fine-tuned model reads better in
context (OCR-text F1 $0.584$--$0.602$, own detection, whole frame) than either
specialist engine granted oracle per-player crops (\cref{tab:main} is the
full-frame tier; Florence-2 OOTB $0.437$, PP-OCRv5
$0.357$). A paired frame-level bootstrap ($10{,}000$ resamples of the $567$
frames, the same resample scoring every system; full table in the supplementary)
makes the headline gaps significant---detection F1 $+0.22$ to $+0.30$,
jersey-digit AER $+0.16$ to $+0.18$, OCR-text F1 $+0.09$ to $+0.11$; every
interval excludes $0$---and we are equally explicit where it does not: our
jersey-F1 edge over the strongest API is $+0.02$ $[-0.02,+0.06]$, and our
digit-AER is statistically indistinguishable from the oracle-boxed cascades'
($+0.00$ to $+0.02$, CIs spanning $0$)---expected, since they are handed
ground-truth boxes yet read only a third of the numbers.

\section{Cross-Domain Transplant: WIDER-Attribute}
\label{sec:wider}

To test whether our results reflect a general recipe rather than one dataset, we
apply the same components---schema-token expansion,
grammar-constrained decoding, frozen towers, teacher-forced
confidence readout, and a crop second pass (applied here as late fusion over all
persons rather than confidence-routed re-query)---to a second, public, human-centered
benchmark: human attribute recognition in unconstrained event scenes on
WIDER-Attribute~\cite{widerattribute} (people across 30 in-the-wild event
categories---parades, ceremonies, meetings, rescues---rather than street
pedestrians; test split: 6\,918 images, 29\,177 person instances, 14 binary
attributes). The
only change is the slot vocabulary (team/jersey/scene become 14 attribute flags);
we fine-tune \emph{only} on the WIDER training split---this is not zero-shot
transfer of the sports weights, but a re-instantiation of the same procedure.
Ground-truth boxes are provided at inference so the number is comparable to the
attribute literature, which assumes given boxes.

\subsubsection{Result.}
The model reaches \textbf{93.1 mAP} given-box (late fusion of a whole-image and a
per-person-crop pass: scene 89.9, crop 91.9, fusion 93.1), at or above a decade of
task-specific WIDER specialists (\cref{tab:wider}), with a single set of weights
serving every inference protocol. Because our detector also localizes people, we
additionally report a \emph{fully end-to-end} number---detect, crop, and classify
without ground-truth boxes---of \textbf{84.5 mAP}, to our knowledge the first
detection-coupled result---no ground-truth boxes at any stage---under the standard
WIDER-Attribute test protocol; the gap to the given-box number is
dominated by an annotation-sparsity artifact (the detector fires on background
people the annotators skipped: unmatched predictions have median box area $0.023$
and $69.5\%$ arise on images with more predictions than labels), so we treat the
end-to-end figure as a lower bound and the given-box figure as the fair
comparison point. The scope of this transplant is stated explicitly: WIDER's
protocol tests the \emph{recipe's} generality---schema, grammar, confidence
readout, second pass---not the text--person association claim itself; the
end-to-end variant exercises grounding but not text binding.

\begin{table}[tbp]
  \caption{WIDER-Attribute in context (\cref{sec:wider}). Prior methods are
    \emph{given-box} only. mAP is the historical WIDER metric; mA (mean
    per-attribute accuracy) is common in recent PAR work.
    \textbf{mAP and mA are different metrics and are not comparable across blocks.}
    The table is for context, not a common-protocol ranking.}
  \label{tab:wider}
  \centering
  \scriptsize
  \setlength{\tabcolsep}{4pt}
  \begin{tabular}{@{}llc@{}}
    \toprule
    Method & Metric & Score \\
    \midrule
    \multicolumn{3}{@{}l}{\emph{mAP era (given-box)}}\\
    R*CNN~\cite{rstarcnn}       & mAP & 80.5 \\
    DHC~\cite{widerattribute}   & mAP & 81.3 \\
    SRN~\cite{srn}              & mAP & 86.2 \\
    DIAA~\cite{diaa}            & mAP & 86.4 \\
    Da-HAR~\cite{dahar}         & mAP & 87.3 \\
    VAC~\cite{vac}              & mAP & 87.5 \\
    \midrule
    \multicolumn{3}{@{}l}{\emph{mA era (given-box)}}\\
    VTB~\cite{vtb}              & mA  & 88.2 \\
    PromptPAR~\cite{promptpar}  & mA  & 92.0 \\
    \midrule
    \multicolumn{3}{@{}l}{\emph{Ours (this work)}}\\
    Ours (given-box)            & mAP & \textbf{93.1} \\
    Ours (end-to-end, first reported) & mAP & 84.5 \\
    \bottomrule
  \end{tabular}
\end{table}

\subsubsection{The same phenomenology.}
All three sports findings replicate. \emph{No measurable tuning gain}: extra
adaptation moves the deployed given-box metric by $<0.1$ point ($+0.02$ mAP),
mirroring the LoRA ladder (\cref{sec:exp-abl}); per-attribute threshold tuning
is likewise unhelpful ($-0.8$ mA). \emph{Usable confidence}: the per-field probabilities are
calibrated here (ECE $0.028$ after fusion) and abstaining on the
least-confident $10\%$ of $\approx$409k (person, attribute) decisions cuts
error from $4.1\%$ to $1.5\%$. \emph{Routing, not filtering}:
\texttt{loc\_conf} routes the re-query benefit to low-confidence detections yet
is a weak filter, and objectness saturates, exactly as on sports
(\cref{sec:exp-conf}; a bonus visibility-detection property of the uncertainty
token is described in the supplementary).

\section{Conclusion}
\label{sec:conclusion}

We presented a single-pass perception engine that makes structure intrinsic to
generation: a 0.77B VLM fine-tuned to emit every per-person attribute inside a
grammar-constrained sequence, so association is a property of decoding, not a
post-hoc stitch. It is schema-valid on every frame, leads detection ($0.95$ F1),
reads at full coverage where the APIs abstain, at far lower latency, and cuts
jersey-digit association error $\approx4\times$ below theirs. In our regime
(frozen towers, a 0.77B backbone, single seed), further parameter-efficient
tuning showed no measurable gain under the tested configurations---once the
grammar is learned, accuracy here is governed by structure---and the identical
recipe on WIDER-Attribute ($93.1$ mAP given-box) reproduces it: a property of
the recipe, not a lucky checkpoint.

\subsubsection{Limitations.}
Accuracy on small players is bounded by legible pixels: the $1000$-bin coordinate
quantization and finite decoder resolution limit tiny-text reading, and quality is
detection-recall-bound, so residual error concentrates in the small-athlete
bucket (\cref{sec:exp-size}), mitigated but not eliminated by the crop re-query.
Confidence thresholds and re-query padding are validation-selected, not learned,
and the raw likelihood is over-confident (ECE $0.12$--$0.15$). Our cascade
baselines are zero-shot / oracle-assisted (ground-truth boxes isolate stitching
from detector quality); the twelve-person cap is non-binding on our data, and the
grammar in practice \emph{reduces} wall-clock $\approx\!7\times$
(\cref{sec:exp-abl}). All runs use a single seed, with tuning deltas gated by the
pre-registered noise floor; residual failures are in the supplementary.
Confidence-gated override of pass~1, temperature scaling, and a domain-trained
cascade comparison are natural next steps, each validation-decidable and left
untuned here.

\subsubsection{Ethics and intended use.}
The system reads jersey numbers and clusters athletes into \emph{relative}
per-frame teams in professional broadcast footage only---no identity or face
recognition---for sports analytics and archive indexing, not surveillance.

\subsubsection*{Disclosure of interests.}
The authors are employed by WSC Sports; they have no other competing interests.

\bibliographystyle{splncs04}
\bibliography{main}

\begin{thebibliography}{10}
\providecommand{\url}[1]{\texttt{#1}}
\providecommand{\urlprefix}{URL }
\providecommand{\doi}[1]{https://doi.org/#1}

\bibitem{detr}
Carion, N., Massa, F., Synnaeve, G., Usunier, N., Kirillov, A., Zagoruyko, S.:
  End-to-end object detection with transformers. In: ECCV (2020)

\bibitem{pix2seq}
Chen, T., Saxena, S., Li, L., Fleet, D.J., Hinton, G.: {Pix2Seq}: A language
  modeling framework for object detection. In: ICLR (2022)

\bibitem{vtb}
Cheng, X., Jia, M., Wang, Q., Zhang, J.: A simple visual-textual baseline for
  pedestrian attribute recognition. IEEE TCSVT  (2022)

\bibitem{genre}
De~Cao, N., Izacard, G., Riedel, S., Petroni, F.: Autoregressive entity
  retrieval. In: ICLR (2021)

\bibitem{davit}
Ding, M., Xiao, B., Codella, N., Luo, P., Wang, J., Yuan, L.: {DaViT}: Dual
  attention vision transformers. In: ECCV (2022)

\bibitem{ppocr}
Du, Y., Li, C., Guo, R., Yin, X., Liu, W., Zhou, J., Bai, Y., Yu, Z., Yang, Y.,
  Dang, Q., Wang, H.: {PP-OCR}: A practical ultra lightweight {OCR} system.
  arXiv:2009.09941 (2020)

\bibitem{selective}
Geifman, Y., El-Yaniv, R.: Selective classification for deep neural networks.
  In: NeurIPS (2017)

\bibitem{gemini}
{Gemini Team, Google}: {Gemini}: A family of highly capable multimodal models.
  arXiv:2312.11805 (2023)

\bibitem{rstarcnn}
Gkioxari, G., Girshick, R., Malik, J.: Contextual action recognition with
  {R*CNN}. In: ICCV (2015)

\bibitem{grad2025jersey}
Grad, {\L}.: Single-stage uncertainty-aware jersey number recognition in
  soccer. In: Proceedings of the IEEE/CVF Conference on Computer Vision and
  Pattern Recognition (CVPR) Workshops. pp. 6148--6156 (2025)

\bibitem{calibration}
Guo, C., Pleiss, G., Sun, Y., Weinberger, K.Q.: On calibration of modern neural
  networks. In: ICML (2017)

\bibitem{vac}
Guo, H., Zheng, K., Fan, X., Yu, H., Wang, S.: Visual attention consistency
  under image transforms for multi-label image classification. In: CVPR (2019)

\bibitem{lora}
Hu, E.J., Shen, Y., Wallis, P., Allen-Zhu, Z., Li, Y., Wang, S., Wang, L.,
  Chen, W.: {LoRA}: Low-rank adaptation of large language models. In: ICLR
  (2022)

\bibitem{llmpar}
Jin, J., Wang, X., Zhu, Q., Wang, H., Li, C.: Pedestrian attribute recognition:
  A new benchmark dataset and a large language model augmented framework. In:
  AAAI (2025)

\bibitem{donut}
Kim, G., Hong, T., Yim, M., Nam, J., Park, J., Yim, J., Hwang, W., Yun, S.,
  Han, D., Park, S.: {OCR}-free document understanding transformer. In: ECCV
  (2022)

\bibitem{sam}
Kirillov, A., Mintun, E., Ravi, N., Mao, H., Rolland, C., Gustafson, L., Xiao,
  T., Whitehead, S., Berg, A.C., Lo, W.Y., Doll{\'a}r, P., Girshick, R.:
  Segment anything. In: ICCV (2023)

\bibitem{koshkina2024jersey}
Koshkina, M., Elder, J.H.: A general framework for jersey number recognition in
  sports video. In: Proceedings of the IEEE/CVF Conference on Computer Vision
  and Pattern Recognition (CVPR) Workshops. pp. 3235--3244 (2024)

\bibitem{trocr}
Li, M., Lv, T., Chen, J., Cui, L., Lu, Y., Florencio, D., Zhang, C., Li, Z.,
  Wei, F.: {TrOCR}: Transformer-based optical character recognition with
  pre-trained models. In: AAAI (2023)

\bibitem{widerattribute}
Li, Y., Huang, C., Loy, C.C., Tang, X.: Human attribute recognition by deep
  hierarchical contexts. In: ECCV (2016)

\bibitem{apr}
Lin, Y., Zheng, L., Zheng, Z., Wu, Y., Hu, Z., Yan, C., Yang, Y.: Improving
  person re-identification by attribute and identity learning. Pattern
  Recognition  (2019)

\bibitem{groundingdino}
Liu, S., Zeng, Z., Ren, T., Li, F., Zhang, H., Yang, J., Li, C., Yang, J., Su,
  H., Zhu, J., Zhang, L.: {Grounding DINO}: Marrying {DINO} with grounded
  pre-training for open-set object detection. In: ECCV (2024)

\bibitem{kosmos2}
Peng, Z., Wang, W., Dong, L., Hao, Y., Huang, S., Ma, S., Wei, F.: {Kosmos-2}:
  Grounding multimodal large language models to the world. In: ICLR (2024)

\bibitem{diaa}
Sarafianos, N., Xu, X., Kakadiaris, I.A.: Deep imbalanced attribute
  classification using visual attention aggregation. In: ECCV (2018)

\bibitem{picard}
Scholak, T., Schucher, N., Bahdanau, D.: {PICARD}: Parsing incrementally for
  constrained auto-regressive decoding from language models. In: Proc. Conf.
  Empirical Methods in Natural Language Processing (EMNLP) (2021)

\bibitem{soccernetgsr}
Somers, V., Joos, V., Cioppa, A., Giancola, S., Ghasemzadeh, S.A., Magera, F.,
  Standaert, B., Mansourian, A.M., Zhou, X., Kasaei, S., Ghanem, B., Alahi, A.,
  Van~Droogenbroeck, M., De~Vleeschouwer, C.: {SoccerNet} game state
  reconstruction: End-to-end athlete tracking and identification on a minimap.
  In: Proceedings of the IEEE/CVF Conference on Computer Vision and Pattern
  Recognition (CVPR) Workshops. pp. 3293--3305 (2024)

\bibitem{letmespeak}
Tam, Z.R., Wu, C.K., Tsai, Y.L., Lin, C.Y., Lee, H.y., Chen, Y.N.: Let me speak
  freely? {A} study on the impact of format restrictions on performance of
  large language models. In: EMNLP Industry Track (2024)

\bibitem{promptpar}
Wang, X., Jin, J., Li, C., Tang, J., Zhang, C., Wang, W.: Pedestrian attribute
  recognition via {CLIP}-based prompt vision-language fusion. IEEE TCSVT
  (2024)

\bibitem{outlines}
Willard, B.T., Louf, R.: Efficient guided generation for large language models.
  arXiv:2307.09702 (2023)

\bibitem{dahar}
Wu, M., Huang, D., Guo, Y., Wang, Y.: Distraction-aware feature learning for
  human attribute recognition via coarse-to-fine attention. In: AAAI (2020)

\bibitem{florence2}
Xiao, B., Wu, H., Xu, W., Dai, X., Hu, H., Lu, Y., Zeng, M., Liu, C., Yuan, L.:
  {Florence-2}: Advancing a unified representation for a variety of vision
  tasks. In: CVPR (2024)

\bibitem{autover}
Xiao, Z., Gong, M., Cascante-Bonilla, P., Zhang, X., Wu, J., Ordonez, V.:
  Grounding language models for visual entity recognition. In: ECCV (2024)

\bibitem{gmner}
Yu, J., Li, Z., Wang, J., Xia, R.: Grounded multimodal named entity recognition
  on social media. In: ACL. pp. 9141--9154 (2023)

\bibitem{srn}
Zhu, F., Li, H., Ouyang, W., Yu, N., Wang, X.: Learning spatial regularization
  with image-level supervisions for multi-label image classification. In: CVPR
  (2017)

\end{thebibliography}
\end{document}